\documentclass[10pt,twocolumn,letterpaper]{article}

\usepackage{cvpr}
\usepackage{times}
\usepackage{epsfig}
\usepackage{graphicx}
\usepackage{amsmath}
\usepackage{amssymb}
\usepackage{algorithm}
\usepackage[noend]{algpseudocode}
\usepackage{booktabs}

\makeatletter
\def\BState{\State\hskip-\ALG@thistlm}
\makeatother

\usepackage[breaklinks=true,bookmarks=true]{hyperref}

\cvprfinalcopy % *** Uncomment this line for the final submission

\def\cvprPaperID{968} % *** Enter the CVPR Paper ID here

\begin{document}

%%%%%%%%% TITLE
\title{The More You Know: Using Knowledge Graphs for Image Classification}

\author{Kenneth Marino, Ruslan Salakhutdinov, Abhinav Gupta \\
Carnegie Mellon University\\
5000 Forbes Ave, Pittsburgh, PA 15213\\
{\tt\small \{kdmarino, rsalakhu, abhinavg\}@cs.cmu.edu}
}

\maketitle
%\thispagestyle{empty}

%%%%%%%%% ABSTRACT
\begin{abstract}
One characteristic that sets humans apart from modern learning-based computer vision algorithms is the ability to acquire knowledge about the world and use that knowledge to reason about the visual world. Humans can learn about the characteristics of objects and the relationships that occur between them to learn a large variety of visual concepts, often with few examples. This paper investigates the use of structured prior knowledge in the form of knowledge graphs and shows that using this knowledge improves performance on image classification. We build on recent work on end-to-end learning on graphs, introducing the Graph Search Neural Network as a way of efficiently incorporating large knowledge graphs into a vision classification pipeline. We show in a number of experiments that our method outperforms standard neural network baselines for multi-label classification.
\end{abstract}

%%%%%%%%% BODY TEXT
\vspace{-.5cm}
\section{Introduction}
Our world contains millions of visual concepts understood by humans. These often are ambiguous (tomatoes can be red or green), overlap (vehicles includes both cars and planes) and have dozens or hundreds of subcategories (thousands of specific kinds of insects). While some visual concepts are very common such as person or car, most categories have many fewer examples, forming a long-tail distribution~\cite{Zhu14}. And yet, even when only shown a few or even one example, humans have the remarkable ability to recognize these categories with high accuracy. In contrast, while modern learning-based approaches can recognize some categories with high accuracy, it usually requires thousands of labeled examples for each of these categories. Given how large, complex and dynamic the space of visual concepts is, this approach of building large datasets for every concept is unscalable. Therefore, we need to ask what humans have that current approaches do not.

\begin{figure}[t]
\begin{center}
   \includegraphics[width=1\linewidth]{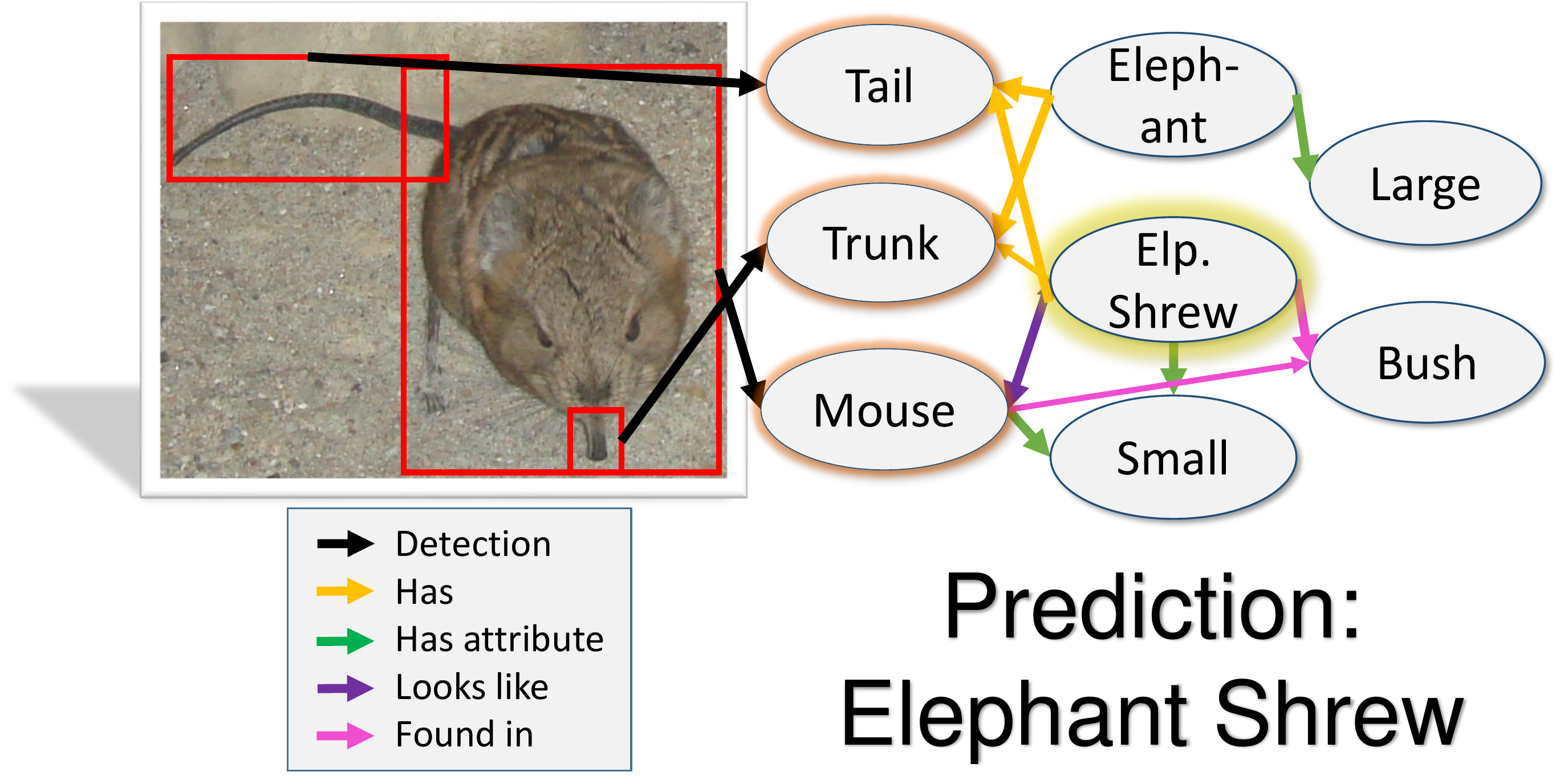}
\end{center}
   \caption{Example of how semantic knowledge about the world aids classification. Here we see an elephant shrew. Humans are able to make the correct classification based on what we know about the elephant shrew and other similar animals.}
\label{fig:teaser}
\vspace{-.5cm}
\end{figure}

One possible answer to this is structured knowledge and reasoning. Humans are not merely appearance-based classifiers; we gain knowledge of the world from experience and language. We use this knowledge in our everyday lives to recognize objects. For instance, we might have read in a book about the ``elephant shrew'' (maybe even seen an example) and will have gained knowledge that is useful for recognizing one. Figure~\ref{fig:teaser} illustrates how we might use our knowledge about the world in this problem. We might know that an elephant shrew looks like a mouse, has a trunk and a tail, is native to Africa, and is often found in bushes. With this information, we could probably identify the elephant shrew if we saw one in the wild. We do this by first recognizing (we see a small mouse-like object with a trunk in a bush), recalling knowledge (we think of animals we have heard of and their parts, habitat, and characteristics) and then reasoning (it is an elephant shrew because it has a trunk and a tail, and looks like a mouse while mice and elephants do not have all these characteristics). With this information, even if we have only seen one or two pictures of this animal, we would be able to classify it.

There has been a lot of work in end-to-end learning on graphs or neural network trained on graphs~\cite{Scarselli09, Bruna13, Duvenaud15, LeCun15, Niepert16, Micheli09, Gori05, DiMassa06}. Most of these approaches either extract features from the graph or they learn a propagation model that transfers evidence between nodes conditional on the type of edge. An example of this is the Gated Graph Neural Network~\cite{Li16} which takes an arbitrary graph as input. Given some initialization specific to the task, it learns how to propagate information and predict the output for every node in the graph. This approach has been shown to solve basic logical tasks as well as program verification. 

Our work improves on this model and adapts end-to-end graph neural networks to multi-label image classification. We introduce the Graph Search Neural Network (GSNN) which uses features from the image to efficiently annotate the graph, select a relevant subset of the input graph and predict outputs on nodes representing visual concepts. These output states are then used to classify the objects in the image. GSNN learns a propagation model which reasons about different types of relationships and concepts to produce outputs on the nodes which are then used for image classification. Our new architecture mitigates the computational issues with the Gated Graph Neural Networks for large graphs which allows our model to be efficiently trained for image tasks using large knowledge graphs. We show how our model is effective at reasoning about concepts to improve image classification tasks. Importantly, our GSNN model is also able to provide explanations on classifications by following how the information is propagated in the graph.

The major contributions of this work are (a) the introduction of the GSNN as a way of incorporating potentially large knowledge graphs into an end-to-end learning system that is computationally feasible for large graphs; (b) a framework for using noisy knowledge graphs for image classification; and (c) the ability to explain our image classifications by using the propagation model. Our method significantly outperforms baselines for multi-label classification. 

\section{Related Work}
\vspace{-.2cm}
Learning knowledge graphs~\cite{Chen13, NELL10, Viske} and using graphs for visual reasoning~\cite{Zhu14, memex} has recently been of interest to the vision community. For reasoning on graphs, several approaches have been studied. For example, \cite{zhu2014eccv} collects a knowledge base and then queries this knowledge base to do first-order probabilistic reasoning to predict affordances. \cite{memex} builds a graph of exemplars for different categories and uses the spatial relationships to perform contextual reasoning. Approaches such as \cite{pra} use random walks on the graphs to learn patterns of edges while performing the walk and predict new edges in the knowledge graph. There has also been some work using a knowledge base for image retrieval~\cite{FeiFei15_2} or answering visual queries~\cite{Zhu15}, but these works are focused on building and then querying knowledge bases rather than using existing knowledge bases as side information for some vision task.

However, none of these approaches have been learned in an end-to-end manner and the propagation model on the graph is mostly hand-crafted. More recently, learning from knowledge graphs using neural networks and other end-to-end learning systems to perform reasoning has become an active area of research. Several works treat graphs as a special case of a convolutional input where, instead of pixel inputs connected to pixels in a grid, we define the inputs as connected by an input graph, relying on either some global graph structure or doing some sort of pre-processing on graph edges ~\cite{Bruna13, Duvenaud15, LeCun15, Niepert16}. However, most of these approaches have been tried on smaller, cleaner graphs such as molecular datasets. In vision problems, these graphs encode contextual and common-sense relationships and are significantly larger and noisier. 

Li and Zemel present Graph Gated Neural Networks (GGNN)~\cite{Li16} which uses neural networks on graph structured data. This paper (an extension of Graph Neural Networks~\cite{Scarselli09}) serves as the foundation for our Graph Search Neural Network (GSNN). Several papers have found success using variants of Graph Neural Networks applied to various simple domains such as quantitative structure-property relationship (QSPR) analysis in chemistry~\cite{Micheli09} and subgraph matching and other graph problems on toy datasets~\cite{Gori05}. GGNN is a fully end-to-end network that takes as input a directed graph and outputs either a classification over the entire graph or an output for each node. For instance, for the problem of graph reachability, GGNN is given a graph, a start node and end node, and the GGNN will have to output whether the end node is reachable from the start node. They show results for logical tasks on graphs and more complex tasks such as program verification. 

There is also a substantial amount of work on various types of kernels defined for graphs~\cite{Vishwanathan10} such as diffusion kernels~\cite{Kondor02}, graphlet kernels~\cite{Shervashidze09}, Weisfeiler-Lehman graph kernels~\cite{Shervashidze11}, deep graph kernels~\cite{Vishwanathan15}, graph invariant kernels~\cite{Orsini15} and shortest-path kernels~\cite{Borgwardt05}. The methods have various ways of exploiting common graph structures, however, these approaches are only helpful for kernel-based approaches such as SVMs which do not compare well with neural network architectures in vision.

%There has been some, but not a lot of work encorporating learning on graphs and images, but not in the manner of this paper. \cite{zhu2014eccv} collects a knowledge base and then queries this knowledge base to do first-order probabilistic reasoning to predict affordances. This system cannot be trained end-to-end and it's unclear how this model would generalize to other vision tasks. There has also been some work using a knowledge base for image retrieval~\cite{FeiFei15_2} or answering visual queries~\cite{Zhu15} but these works are focused on building and then querying knowledge bases rather than using existing knowledge bases as side information for some vision task.

Our work is also related to attribute approaches~\cite{Farhadi09} to vision such as \cite{Lampert14} which uses a fixed set of binary attributes to do zero-shot prediction, \cite{Shrivastava14} which uses attributes shared across categories to prevent semantic drift in semi-supervised learning and \cite{Kun12} which automatically discovers attributes and uses them for fine-grained classification. Our work also uses attribute relationships that appear in our knowledge graphs, but also uses relationships between objects and reasons directly on graphs rather than using object-attribute pairs directly.

\section{Methodology}
\subsection{Graph Gated Neural Network}
\vspace{-.2cm}
The idea of GGNN is that given a graph with $N$ nodes, we want to produce some output which can either be an output for every graph node $o_1, o_2, ... o_N$ or a global output $o_G$. This is done by learning a propagation model similar to an LSTM. For each node in the graph $v$, we have a hidden state representation $h_v^{(t)}$ at every time step $t$. We start at $t=0$ with initial hidden states $x_v$ that depends on the problem. For instance, for learning graph reachability, this might be a two bit vector that indicates whether a node is the source or destination node. In case of visual knowledge graph reasoning, $x_v$ can be a one bit activation representing the confidence of a category being present based on an object detector or classifier. 

Next, we use the structure of our graph, encoded in a matrix $A$ which serves to retrieve the hidden states of adjacent nodes based on the edge types between them. The hidden states are then updated by a gated update module similar to an LSTM. The basic recurrence for this \textbf{propagation network} is
\vspace{-.25cm}
\begin{equation}
\label{rec_1}
h_v^{(1)} = [x_v^T, 0]^T
\end{equation}
\vspace{-.55cm}
\begin{equation}
\label{rec_2}
a_v^{(t)} = A_v^T [h_1^{(t-1)} ... h_N^{(t-1)}]^T + b
\end{equation}
\vspace{-.55cm}
\begin{equation}
\label{rec_3}
z_v^t = \sigma(W^z a_v^{(t)}+U^z h_v^{(t-1)})
\end{equation}
\vspace{-.55cm}
\begin{equation}
\label{rec_4}
r_v^t = \sigma(W^r a_v^{(t)}+U^r h_v^{(t-1)})
\end{equation}
\vspace{-.55cm}
\begin{equation}
\label{rec_5}
\widetilde{h_v^t} = tanh(W a_v^{(t)}+U(r_v^t \odot h_v^{(t-1)}))
\end{equation}
\vspace{-.55cm}
\begin{equation}
\label{rec_6}
h_v^{(t)} = (1-z_v^t) \odot h_v^{(t-1)} + z_v^t \odot \widetilde{h_v^t}
\end{equation}
where $h_v^{(t)}$ is the hidden state for node $v$ at time step $t$, $x_v$ is the problem specific annotation, $A_v$ is the adjacency matrix of the graph for node $v$, and $W$ and $U$ are learned parameters. Eq~\ref{rec_1} is the initialization of the hidden state with $x_v$ and empty dimensions. Eq~\ref{rec_2} shows the propagation updates from adjacent nodes. Eq (3-6) combine the information from adjacent nodes and current hidden state of the nodes to compute the next hidden state.  

After $T$ time steps, we have our final hidden states. The node level outputs can then just be computed as
\vspace{-.2cm}
\begin{equation}
\label{output_eq}
o_v = g(h_v^{(T)}, x_v) 
\vspace{-.2cm}
\end{equation}
where $g$ is a fully connected network, the \textbf{output network}, and $x_v$ is the original annotation for the node.

\subsection{Graph Search Neural Network}
The biggest problem in adapting GGNN for image tasks is computational scalability. NEIL~\cite{Chen13} for example has over 2000 concepts, and NELL~\cite{NELL10} has over 2M confident beliefs. Even after pruning to our task, these graphs would still be huge. Forward propagation on the standard GGNN is $O(N^2)$ to the number of nodes $N$ and backward propagation is $O(N^T)$ where $T$ is the number of propagation steps. We perform simple experiments on GGNNs on synthetic graphs and find that after more than about 500 nodes, a forward and backward pass takes over 1 second on a single instance, even when making generous parameter assumptions. On 2,000 nodes, it takes well over a minute for a single image. Using GGNN out of the box is infeasible. 

Our solution to this problem is the Graph Search Neural Network (GSNN). As the name might imply, 
the idea is that rather than performing our recurrent update over all of the nodes of the graph at once, we start with some initial nodes based on our input and only choose to expand nodes which are useful for the final output. Thus, we only compute the update steps over a subset of the graph. So how do we select which subset of nodes to initialize the graph with? During training and testing, we determine initial nodes in the graph based on likelihood of the concept being present as determined by an object detector or classifier. For our experiments, we use Faster R-CNN~\cite{renNIPS15fasterrcnn} for each of the 80 COCO categories. For scores over some chosen threshold, we choose the corresponding nodes in the graph as our initial set of active nodes.

Once we have initial nodes, we also add the nodes adjacent to the initial nodes to the active set. Given our initial nodes, we want to first propagate the beliefs about our initial nodes to all of the adjacent nodes. After the first time step, however, we need a way of deciding which nodes to expand next. We therefore learn a per-node scoring function that estimates how ``important'' that node is. After each propagation step, for every node in our current graph, we predict an importance score
\begin{equation}
\label{importance}
i_v^{(t)} = g_i(h_v, x_v)
\end{equation}
where $g_i$ is a learned network, the \textbf{importance network}. 

\begin{figure}[t]
\begin{center}
   \includegraphics[width=1\linewidth]{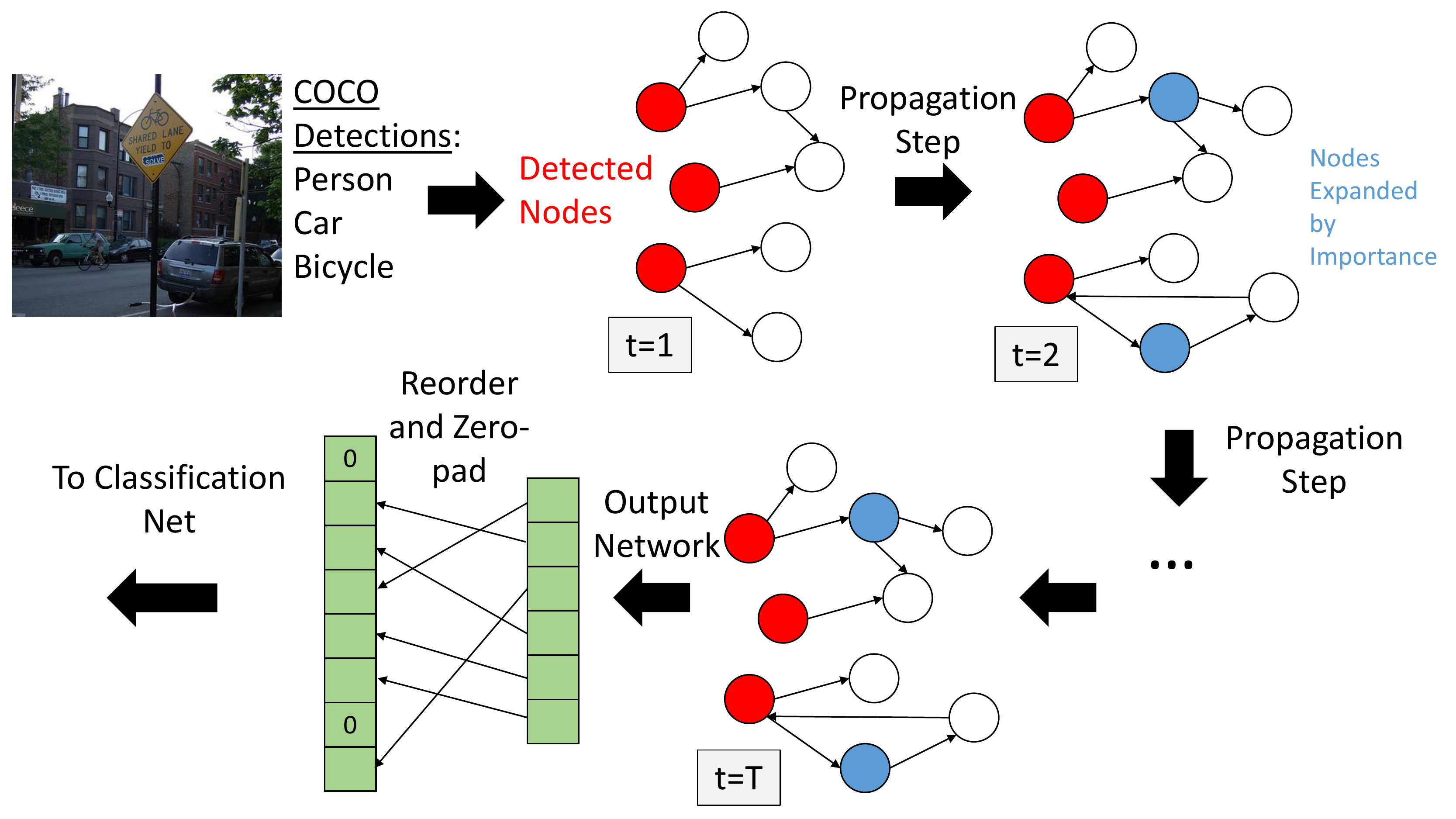}
\end{center}
   \caption{Graph Search Neural Network expansion. Starts with detected nodes and expands neighbors. Adds nodes adjacent to expand nodes predicted by importance net.}
   \vspace{-.5cm}
\label{fig:expand}
\end{figure}

Once we have values of $i_v$, we take the top $P$ scoring nodes that have never been expanded and add them to our expanded set, and add all nodes adjacent to those nodes to our active set. Figure~\ref{fig:expand} illustrates this expansion. At $t=1$ only the detected nodes are expanded. At $t=2$ we expand chosen nodes based on importance values and add their neighbors to the graph. At the final time step $T$ we compute the per-node-output and re-order and zero-pad the outputs into the final classification net.

To train the importance net, we assign target importance value to each node in the graph for a given image. Nodes corresponding to ground-truth concepts in an image are assigned an importance value of 1. The neighbors of these nodes are assigned  a value of $\gamma$. Nodes which are two-hop away have value $\gamma^2$ and so on. The idea is that nodes closest to the final output are the most important to expand.

\begin{figure}[t]
\begin{center}
   \includegraphics[width=1\linewidth]{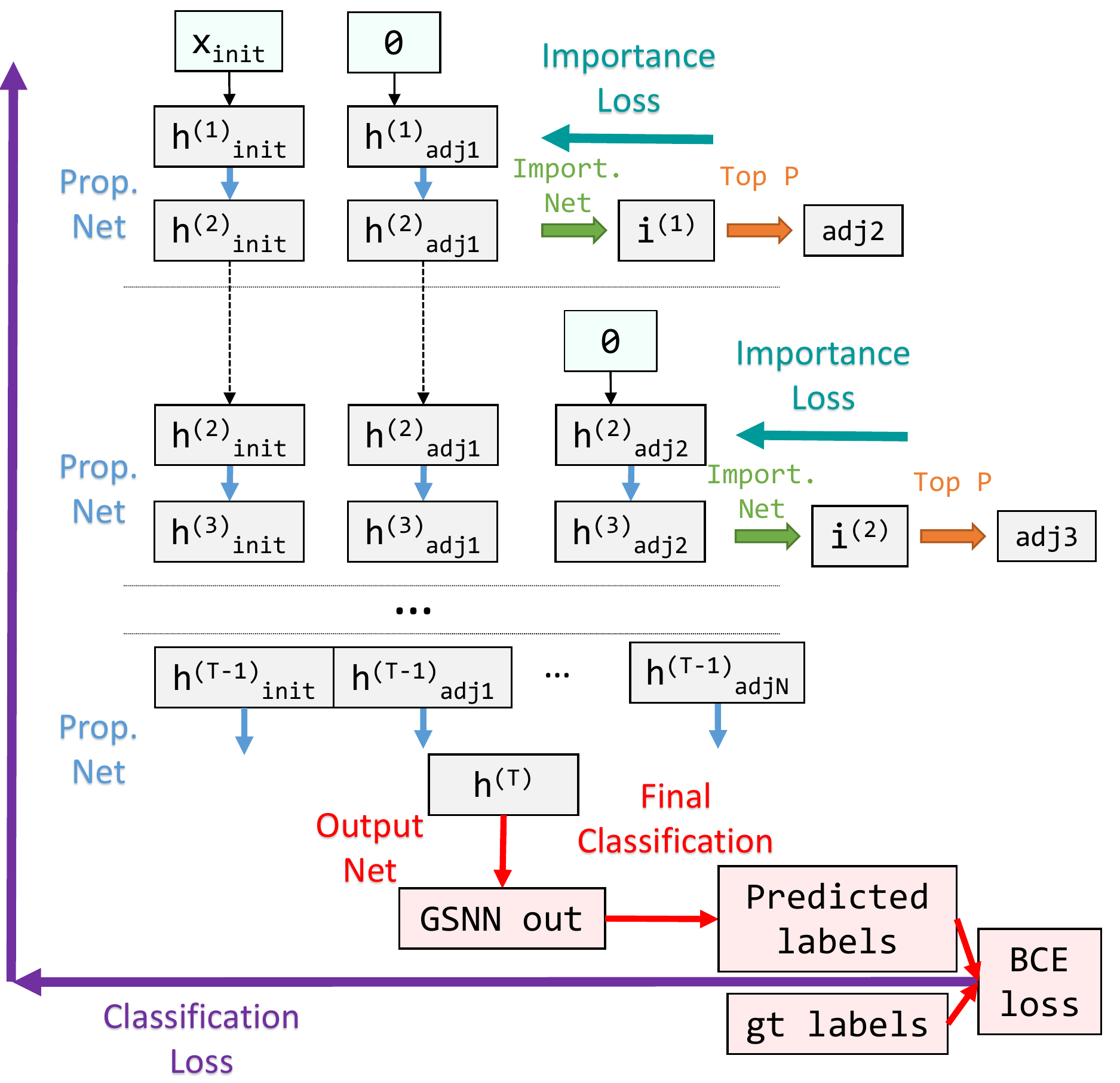}
\end{center}
   \caption{Graph Search Neural Network diagram. Shows initialization of hidden states, addition of new nodes as graph is expanded and the flow of losses through the output, propagation and importance nets. }
   \vspace{-.5cm}
\label{fig:net}
\end{figure}

We now  have an end-to-end network which takes as input a set of initial nodes and annotations and outputs a per-node output for each of the active nodes in the graph. It consists of three sets of networks: the propagation net, the importance net, and the output net. The final loss from the image problem can be backpropagated from the final output of the pipeline back through the output net and the importance loss is backpropagated through each of the importance outputs. See Figure~\ref{fig:net} to see the GSNN architecture. First $x_{init}$, the detection confidences initialize $h^{(1)}_{init}$, the hidden states of the initially detected nodes. We then initialize $h^{(1)}_{adj1}$, the hidden states of the adjacent nodes, with $0$. We then update the hidden states using the propagation net. The values of $h^{(2)}$ are then used to predict the importance scores $i^{(1)}$, which are used to pick the next nodes to add $adj2$. These nodes are then initialized with $h^{(2)}_{adj2}=0$ and the hidden states are updated again through the propagation net. After $T$ steps, we then take all of the accumulated hidden states $h^{T}$ to predict the GSNN outputs for all the active nodes. During backpropagation, the binary cross entropy (BCE) loss is fed backward through the output layer, and the importance losses are fed through the importance networks to update the network parameters.

One final detail is the addition of a ``node bias'' into GSNN. In GGNN, the per-node output function $g(h^{(T)}_v, x_v)$ takes in the hidden state and initial annotation of the node~$v$ to compute its output. In a certain sense it is agnostic to the meaning of the node. That is, at train or test time, GSNN takes in a graph it has perhaps never seen before, and some initial annotations $x_v$ for each node. It then uses the structure of the graph to propagate those annotations through the network and then compute an output. The nodes of the graph could have represented anything from human relationships to a computer program. However, in our graph network, the fact that a particular node represents ``horse'' or ``cat'' will probably be relevant, and we can also constrain ourselves to a static graph over image concepts. Hence we introduce node bias terms that, for every node in our graph, has some learned values. Our output equations are now $g(h^{(T)}_v, x_v, n_v)$ where $n_v$ is a bias term that is tied to a particular node $v$ in the overall graph. This value is stored in a table and its value are updated by backpropagation.

\subsection{Image pipeline and baselines}
Another problem we face adapting graph networks for vision problems is how to incorporate the graph network into an image pipeline. For classification, this is fairly straightforward. We take the output of the graph network, reorder it so that nodes always appear in the same order into the final network, and zero pad any nodes that were not expanded. Therefore, if we have a graph with $316$ node outputs, and each node predicts a $5$-dim hidden variable, we create a $1580$-dim feature vector from the graph. We also concatenate this feature vector with fc7 layer ($4096$-dim) of a fine-tuned VGG-16 network~\cite{VGG14} and top-score for each COCO category predicted by Faster R-CNN ($80$-dim). This $5756$-dim feature vector is then fed into 1-layer final classification network trained with dropout.

For baselines, we compare to: (1) VGG Baseline - feed just fc7 into final classification net; (2) Detection Baseline - feed fc7 and top COCO scores into final classification net. 

\section{Results}
\subsection{Datasets}
For our experiments, we wanted to test on a dataset that represents the complex, noisy visual world with its many different kinds of objects, where labels are potentially ambiguous and overlapping, and categories fall into a long-tail distribution~\cite{Zhu14}. Humans do well in this setting, but vision algorithms still struggle with it. To this end, we chose the Visual Genome dataset~\cite{krishnavisualgenome} v1.0. 

Visual Genome contains over 100,000 natural images from the Internet. Each image is labeled with objects, attributes and relationships between objects entered by human annotators. Annotators could enter any object in the image rather than from a predefined list, so as a result there are thousands of object labels with some being more common and most having many fewer examples. There are on average 21 labeled objects in an image, so compared to datasets such as ImageNet~\cite{ILSVRC15} or PASCAL~\cite{pascal-voc-2012}, the scenes we are considering are far more complex. Visual Genome is also labeled with object-object relationships and object-attribute relationships which we use for GSNN.

In our experiments, we create a subset from Visual Genome which we call Visual Genome multi-label dataset or VGML. In VGML, we take the 200 most common objects in the dataset and the 100 most common attributes and also add any COCO categories not in those 300 for a total of 316 visual concepts. Our task is then multi-label classification: for each image predict which subset of the 316 total categories appear in the scene. We randomly split the images into a roughly 80-20 train/test split. Since we used pre-trained detectors from COCO, we ensure none of our test images overlap with our detector's training images.

We also evaluate out method on the more standard COCO dataset~\cite{LinMBHPRDZ14} to show that our approach is useful on multiple datasets and that our method does not rely on graphs built specifically for our datasets. We train and test in the multi-label setting~\cite{MisraNoisy16}, and evaluate on the minival set~\cite{renNIPS15fasterrcnn}.

\subsection{Building the Knowledge Graph}
We also use Visual Genome as a source for our knowledge graph. Using only the train split, we build a knowledge graph connecting the concepts using the most common object-attribute and object-object relationships in the dataset. Specifically, we counted how often an object/object relationship or object/attribute pair occurred in the training set, and pruned any edges that had fewer than 200 instances. This leaves us with a graph over all of the images with each edge being a common relationship. The idea is that we would get very common relationships (such as grass is green or person wears clothes) but not relationships that are rare and only occur in single images (such as person rides zebra).

The Visual Genome graphs are useful for our problem because they contain scene-level relationships between objects, e.g. person wears pants or fire hydrant is red and thus allow the graph network to reason about what is in a scene. However, it does not contain useful semantic relationships. For instance, it might be helpful to know that dog is an animal if our visual system sees a dog and one of our labels is animal. To address this, we also create a version of graph by fusing the Visual Genome Graphs with WordNet~\cite{Miller95}. Using the subset of WordNet from~\cite{gu2015traversing}, we first collect new nodes in WordNet not in our output label by including those which directly connect to our output labels and thus likely to be relevant and add them to a combined graph. We then take all of the WordNet edges between these nodes and add them to our combined graph.

\subsection{Training details}
We jointly train all parts of the pipeline (except for the detectors). All models are trained with Stochastic Gradient Descent, except GSNN which is trained using ADAM~\cite{Kingma15}. We use an initial learning rate of $0.05$, $0.005$ for the VGG net before $fc7$, decreasing by a factor of $0.1$ every $10$ epochs, an L2 penalty of $1e^{-6}$ and a momentum of $0.5$. We set our GSNN hidden state size to $10$,  importance discount factor $\gamma$ to $0.3$, number of time steps $T$ to $3$, initial confidence threshold to $0.5$ and our expand number $P$ to $5$. Our GSNN importance and output networks are single layer networks with sigmoid activations. All networks were trained for $20$ epochs with a batch size of 16. 

\subsection{Quantitative Evaluation}
Table~\ref{table:VG} shows the result of our method on Visual Genome multi-label classification. In this experiment, the combined Visual Genome, WordNet graph outperforms the Visual Genome graph. This suggests that including the outside semantic knowledge from WordNet and performing explicit reasoning on a knowledge graph allows our model to learn better representations compared to the other models.

\begin{figure}[t]
\begin{center}
   \includegraphics[width=1\linewidth]{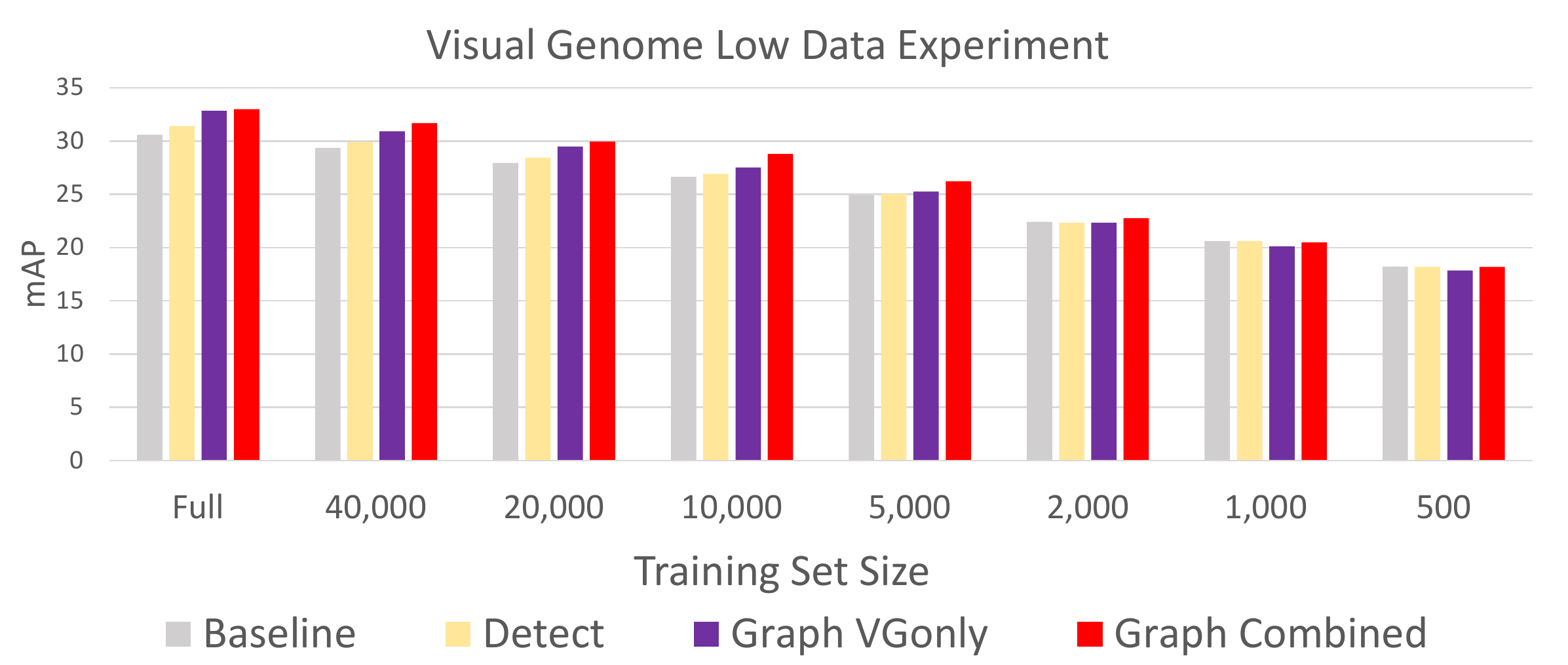}
\end{center}
\vspace{-.4cm}
\caption{Mean Average Precision on Visual Genome in the low data setting. Shows performance for all methods for the full dataset, 40,000, 20,000, 10,000, 5,000, 2,000, 1,000, and 500 training examples.}
\label{fig:VGlowdata}
\end{figure}

We also perform experiments to test the effect of limiting the size of the training dataset has on performance. Figure~\ref{fig:VGlowdata} shows the results of this experiment on Visual Genome, varying the training set size from the entire training set (approximately 80,000), all the way down to 500 examples. Choosing the subsets of examples for these experiments is done randomly, but each training set is a subset of the larger ones---e.g. all of the examples in the 1,000 set are also in the 2,000 set. We see that, until the 1,000 sample set, the GSNN-based methods all outperform baselines. At 1,000 and 500 examples, all of the methods perform equally. Given the long-tail nature of Visual Genome, it is likely that for fewer than 2,000 samples, many categories do not have enough examples for any method to learn well. This experiment indicates that our method is able to improve even in the low-data case up to a point.

In Table~\ref{table:COCO}, we show results on the COCO multi-label dataset. We can see that the boost from using graph knowledge is more significant than it was on Visual Genome. One possible explanation is that the Visual Genome knowledge graph provides significant information which helps improve the performance on the COCO dataset itself. In the previous Visual Genome experiment, much of the graph information is contained in the labels and images themselves. One other interesting result is that the Visual Genome graph outperforms the combined graph for COCO, though both outperform baselines. One possible reason is that the original VGML graph is smaller, cleaner, and contains more relevant information than the combined graph. Furthermore, in the VGML experiment, WordNet is new outside information for the algorithm helping boost the performance.

One possible concern is the over dependence of the graph reasoning on the set of 80 COCO detectors and initial detections. Therefore, we performed an ablation experiment to see how sensitive our method is to having all of the initial detections. We reran the COCO experiments with both graphs using two different subsets of COCO detectors. The first subset is just the even COCO categories and the second subset is just the odd categories. We see from Table~\ref{table:EvenOdd} that GSNN methods again outperform the baselines.

\begin{table}[t]
\caption{Mean Average Precision for multi-label classification on Visual Genome Multi-Label dataset. Numbers for VGG baseline, VGG baseline with detections, GSNN using Visual Genome graph and GSNN using a combined Visual Genome and WordNet graph.}
\begin{center}
\begin{tabular}{@{}ll@{}}
\toprule
Method & mAP \\ \midrule
VGG & 30.57 \\
VGG+Det & 31.4 \\
GSNN-VG& 32.83 \\
GSNN-VG+WN& \textbf{33}\\ \bottomrule
\end{tabular}
\end{center}
\label{table:VG}
\vspace{-.4cm}
\end{table}

\begin{table}[t]
\caption{Mean Average Precision for multi-label classification on COCO. Numbers for VGG baseline, VGG baseline with detections, GSNN using Visual Genome graph and GSNN using combined Visual Genome and WordNet graph.}
\begin{center}
\begin{tabular}{@{}ll@{}}
\toprule
Method & mAP \\ \midrule
VGG & 69.86\\
VGG+Det & 73.93 \\
GSNN-VG& \textbf{77.57} \\
GSNN-VG+WN& 75.73\\ \bottomrule
\end{tabular}
\end{center}
\label{table:COCO}
\vspace{-.4cm}
\end{table}

\begin{table}[t]
\caption{Mean Average Precision for multi-label classification on COCO, using only odd and even detectors.}
\begin{center}
\begin{tabular}{@{}lll@{}}
\toprule
Method & even mAP & odd mAP \\ \midrule
VGG+Det & 71.87 & 71.73 \\
GSNN-VG & 73 & 73.43 \\
GSNN-VG+WN & \textbf{73.59} & \textbf{73.97} \\ \bottomrule
\end{tabular}
\end{center}
\label{table:EvenOdd}
\vspace{-.8cm}
\end{table}

\begin{figure*}[t]
\begin{center}
   \includegraphics[width=0.9\linewidth]{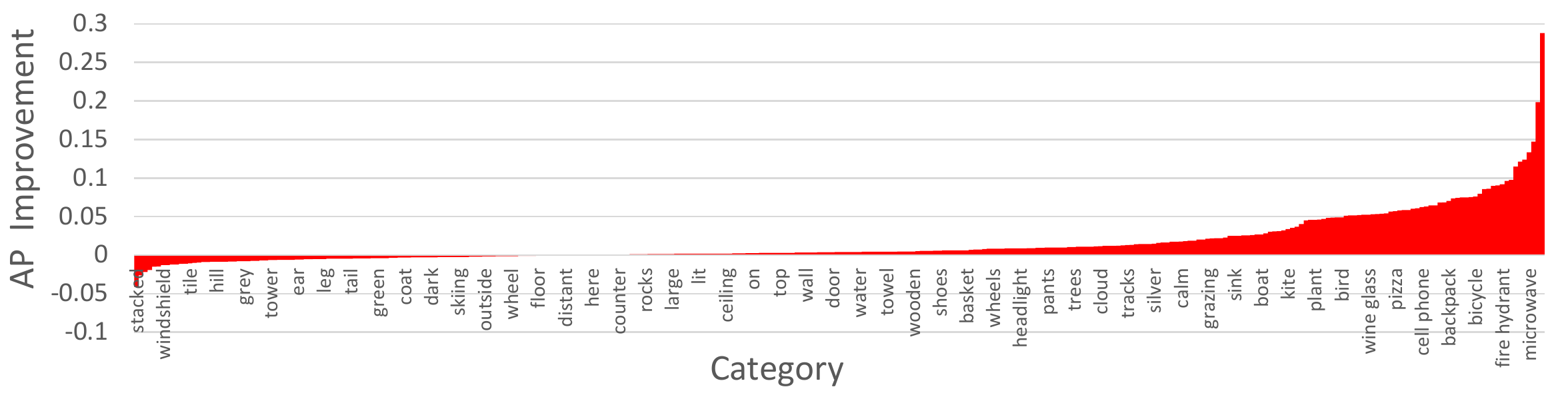}
\end{center}
\vspace{-.4cm}
\caption{Difference in Average Precision for each of the 316 labels in VGML between our GSNN combined graph model and detection baseline for the Visual Genome experiment. Top categories: scissors, donut, frisbee, microwave, fork. Bottom categories: stacked, tiled, light brown, ocean, grassy.}
\label{fig:cats_vg}
\end{figure*}

\begin{figure*}[t]
\begin{center}
   \includegraphics[width=0.9\linewidth]{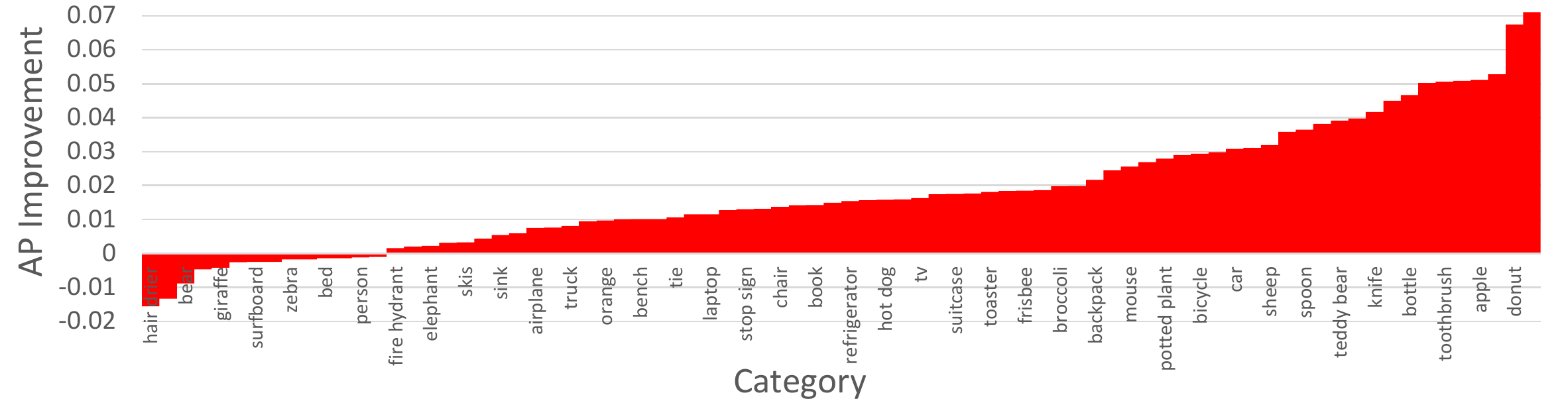}
\end{center}
\vspace{-.4cm}
\caption{Difference in Average Precision for each of the 80 labels in COCO between our GSNN VG graph model and detection baseline for the COCO experiment. Top categories: fork, donut, cup, apple, microwave. Bottom categories: hairdryer, parking meter, bear, kite, and giraffe.}
\label{fig:cats_coco}
\vspace{-.4cm}
\end{figure*}

As one might suspect, our method does not perform uniformly on all categories, but rather does better on some categories and worse on others. Figure~\ref{fig:cats_vg} shows the differences in average precision for each category between our GSNN model with the combined graph and the detection baseline for the VGML experiment. Figure~\ref{fig:cats_coco} shows the same for our COCO experiment. Performance on some classes improves greatly, such as ``fork'' in our VGML experiment and ``scissors'' in our COCO experiment. These and other good results on ``knife'' and ``toothbrush'' seem to indicate that the graph reasoning helps especially with small objects in the image. In the next section, we analyze our GSNN models on several examples to try to gain a better intuition as to what the GSNN model is doing and why it does well or poorly on certain examples.

\subsection{Qualitative Evaluation}
\begin{figure*}
\vspace{1cm}
\begin{center}
   \includegraphics[width=1.0\linewidth]{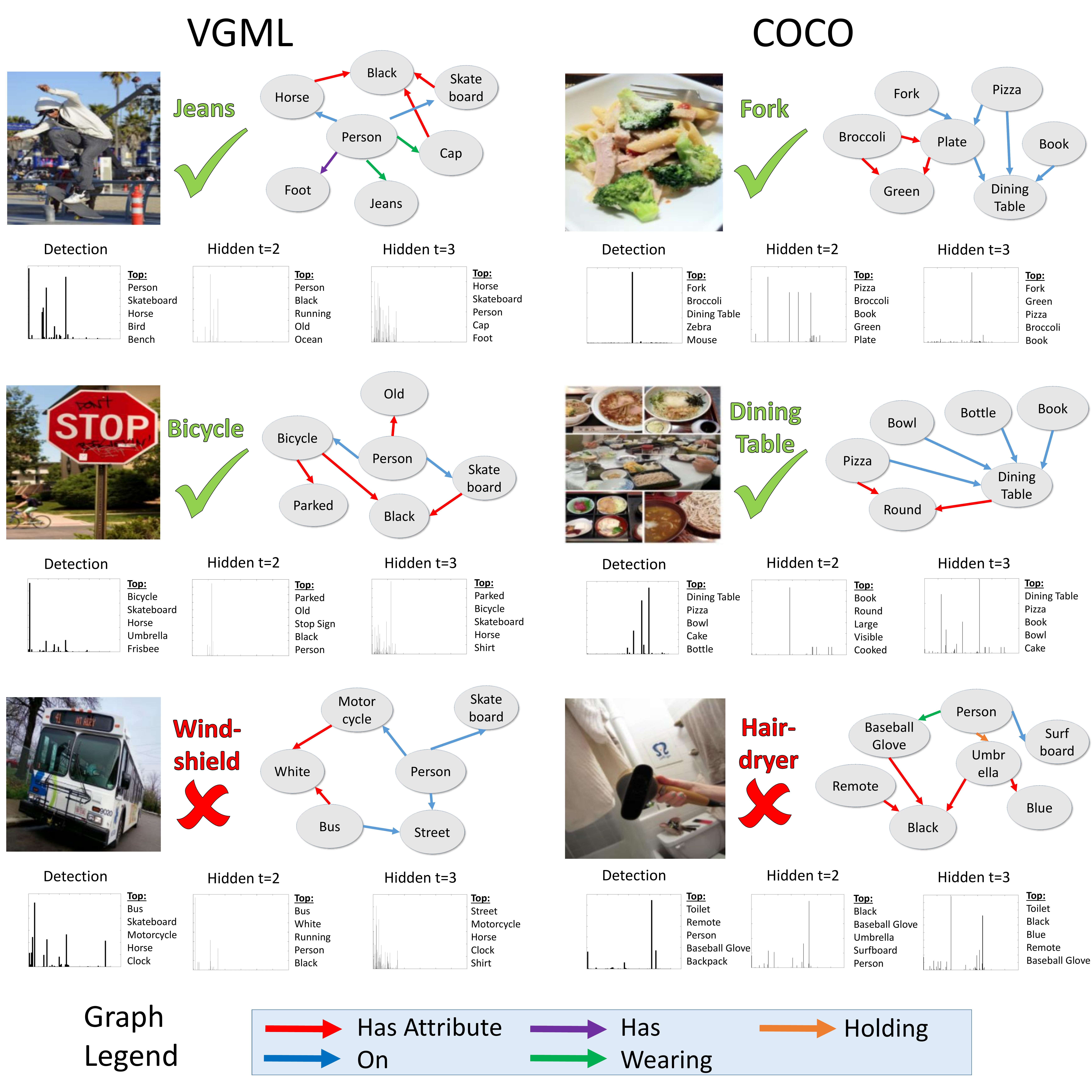}
\end{center}
   \caption{Sensitivity analysis of GSNN in VGML experiment (left) and COCO experiment (right) with the combined graph and Visual Genome graphs respectively. Each example shows the image, part of the knowledge graph expanded during the classification, and the sensitivity values of the initial detections, and the hidden states at time steps 2 and 3 with respect to the output class listed. The top detections and hidden state nodes are printed for convenience since the x-axis is too large to list every class. The top and middle rows show the results for images and classes where the GSNN significantly outperforms the detection baseline to get an intuition for when our method is working. The bottom row shows images and classes where GSNN does worse than the detection baseline to get an idea of when our method fails and why.}
\vspace{1cm}
\label{fig:Sensitivity}
\end{figure*}

One way to analyse the GSNN is to look at the sensitivities of parameters in our model with respect to a particular output. Given a single image $I$, and a single label of interest $y_i$ that appears in the image, we would like to know how information travels through the GSNN and what nodes and edges it uses. We examined the sensitivity of the output to hidden states and detections by computing the partial derivatives $\frac{\partial y_i}{\partial h^{(1)}}$ $\frac{\partial y_i}{\partial h^{(2)}}$ $\frac{\partial y_i}{\partial x_{det}}$ with respect to the category of interest. These values tell us how a small change in the hidden state of a particular node affects a particular output. We would expect to see, for instance, that for labeling elephant, we see a high sensitivity for the hidden states corresponding to grey and trunk.

In this section, we show the sensitivity analysis for the GSNN combined graph model on the VGML experiment and the Visual Genome graph on the COCO experiments. In particular, we examine some classes that performed well under GSNN compared to the detection baseline and a few that performed poorly to try to get a better intuition into why some categories improve more.

Figure~\ref{fig:Sensitivity} shows the graph sensitivity analysis for the experiments with VGML on the left and COCO on the right, showing four examples where GSNN does better and two where it does worse. Each example shows the image, the ground truth output we are analyzing and the sensitivities of the concept of interest with respect to the hidden states of the graph or detections. For convenience, we display the names of the top detections or hidden states. We also show part of the graph that was expanded, to see what relationships GSNN was using.

For the VGML experiment, the top left of Figure~\ref{fig:Sensitivity} shows that using the detection for person, GSNN is able to reason that jeans are more likely since jeans are usually on people in images using the ``wearing'' edge. It is also sensitive to skateboard and horse, and each of these has a second order connection to jeans through person, so it is likely able to capture the fact that people tend to wear jeans while on horses and skateboards. Note that the sensitivities are not the same as the actual detections, so it is not contradictory that horse has high sensitivity. The second row on the left shows a successful example for bicycle, using detections from person and skateboard and the fact that people tend to be ``on'' bicycles and skateboards. The last row shows a failure case for windshield. It correctly correlates with bus, but because the knowledge graph lacks a connection between bus and windshield, the graph network is unable to do better than the detection baseline. On the right, for the COCO experiment, the top example shows that fork is highly correlated with the detection for fork, which should not be surprising. However, it is able to reinforce this detection with the connections between broccoli and dining table, which are both two step connections to fork on the graph. Similarly, the middle example shows that the graph connections for pizza, bowl, and bottle being ``on'' dining table reinforce the detection of dining table. The bottom right shows another failure case. It is able to get the connection between the detection for toilet and hair dryer (both found in the bathroom), but the lack of good connections in the graph prevent the GSNN from improving over the baseline.

\section{Conclusion}
In this paper, we present the Graph Search Neural Network (GSNN) as a way of efficiently using knowledge graphs as extra information to improve image classification. We provide analysis that examines the flow of information through the GSNN and provides insights into why our model improves performance. We hope that this work provides a step towards bringing symbolic reasoning into traditional feed-forward computer vision frameworks.

The GSNN and the framework we use for vision problems is completely general. Our next steps will be to apply the GSNN to other vision tasks, such as detection, Visual Question Answering, and image captioning. Another interesting direction would be to combine the procedure of this work with a system such as NEIL~\cite{Chen13} to create a system which builds knowledge graphs and then prunes them to get a more accurate, useful graph for image tasks.

\noindent {\footnotesize {\bf Acknowledgements}: We would like to thank everyone who took time to review this work and provide helpful comments. This research is based upon work supported in part by the Office of the Director of National Intelligence (ODNI), Intelligence Advanced Research Projects Activity (IARPA). The views and conclusions contained herein are those of the authors and should not be interpreted as necessarily representing the official policies, either expressed or implied of ODNI, IARPA, or the US government. The US Government is authorized to reproduce and distribute the reprints for governmental purposed notwithstanding any copyright annotation therein. This material is based upon work supported by the National Science Foundation Graduate Research Fellowship under Grant No. DGE-1252522 and ONR MURI N000141612007.}
{\small
\bibliographystyle{ieee}
\bibliography{egbib}

\begin{thebibliography}{10}\itemsep=-1pt

\bibitem{Borgwardt05}
K.~M. Borgwardt and H.-P. Kriegel.
\newblock Shortest-path kernels on graphs.
\newblock {\em ICDM}, 2005.

\bibitem{Bruna13}
J.~Bruna, W.~Zaremba, A.~Szlam, and Y.~LeCun.
\newblock Spectral networks and locally connected networks on graphs.
\newblock {\em arXiv preprint arXiv:1312.6203}, 2013.

\bibitem{NELL10}
A.~Carlson, J.~Betteridge, B.~Kisiel, B.~Settles, E.~R. Hruschka, and T.~M.
  Mitchell.
\newblock Toward an architecture for never-ending language learning.
\newblock {\em AAAI}, 2010.

\bibitem{Chen13}
X.~Chen, A.~Shrivastava, and A.~Gupta.
\newblock Neil: Extracting visual knowledge from web data.
\newblock {\em CVPR}, 2013.

\bibitem{Kun12}
K.~Duan, D.~Parikh, D.~Crandall, and K.~Grauman.
\newblock Discovering localized attributes for fine-grained recognition.
\newblock {\em CVPR}, 2012.

\bibitem{Duvenaud15}
D.~K. Duvenaud, D.~Maclaurin, J.~Iparraguirre, R.~Bombarell, T.~Hirzel,
  A.~Aspuru-Guzik, and R.~P. Adams.
\newblock Convolutional networks on graphs for learning molecular fingerprints.
\newblock {\em NIPS}, 2015.

\bibitem{pascal-voc-2012}
M.~Everingham, L.~Van~Gool, C.~K.~I. Williams, J.~Winn, and A.~Zisserman.
\newblock The {PASCAL} {V}isual {O}bject {C}lasses {C}hallenge 2012 {(VOC2012)}
  {R}esults.
\newblock
  http://www.pascal-network.org/challenges/VOC/voc2012/workshop/index.html.

\bibitem{Farhadi09}
A.~Farhadi, I.~Endres, D.~Hoiem, and D.~Forsyth.
\newblock Describing objects by their attributes.
\newblock {\em CVPR}, 2009.

\bibitem{Gori05}
M.~Gori, G.~Monfardini, and F.~Scarselli.
\newblock A new model for learning in graph domains.
\newblock {\em IEEE International Joint Conference on Neural Networks}, 2,
  2005.

\bibitem{gu2015traversing}
K.~Guu, J.~Miller, and P.~Liang.
\newblock Traversing knowledge graphs in vector space.
\newblock In {\em Empirical Methods in Natural Language Processing (EMNLP)},
  2015.

\bibitem{LeCun15}
M.~Henaff, J.~Bruna, and Y.~LeCun.
\newblock Deep convolutional networks on graph-structured data.
\newblock {\em arXiv preprint arXiv:1506.05163}, 2015.

\bibitem{FeiFei15_2}
J.~Johnson, R.~Krishna, M.~Stark, L.-J. Li, D.~A. Shamma, M.~S. Bernstein, and
  L.~Fei-Fei.
\newblock Image retrieval using scene graphs.
\newblock {\em CVPR}, 2015.

\bibitem{Kingma15}
D.~P. Kingma and J.~L. Ba.
\newblock Adam: A method for stochastic optimization.
\newblock {\em ICLR}, 2015.

\bibitem{Kondor02}
R.~I. Kondor and J.~Lafferty.
\newblock Diffusion kernels on graphs and other discrete input spaces.
\newblock {\em ICML}, 2, 2002.

\bibitem{krishnavisualgenome}
R.~Krishna, Y.~Zhu, O.~Groth, J.~Johnson, K.~Hata, J.~Kravitz, S.~Chen,
  Y.~Kalantidis, L.-J. Li, D.~A. Shamma, M.~Bernstein, and L.~Fei-Fei.
\newblock Visual genome: Connecting language and vision using crowdsourced
  dense image annotations.
\newblock 2016.

\bibitem{Lampert14}
C.~H. Lampert, H.~Nickisch, and S.~Harmeling.
\newblock Attribute-based classification for zero-shot visual object
  categorization.
\newblock {\em TPAMI}, 2014.

\bibitem{pra}
N.~Lao, T.~Mitchell, and W.~W. Cohen.
\newblock Random walk inference and learning in a large scale knowledge base.
\newblock {\em NIPS}, 2011.

\bibitem{Li16}
Y.~Li and R.~Zemel.
\newblock Gated graph sequence neural networks.
\newblock {\em ICLR}, 2016.

\bibitem{LinMBHPRDZ14}
T.~Lin, M.~Maire, S.~J. Belongie, R.~B. Girshick, J.~Hays, P.~Perona,
  D.~Ramanan, P.~Doll{\'{a}}r, and C.~L. Zitnick.
\newblock Microsoft {COCO:} common objects in context.
\newblock {\em ECCV}, 2014.

\bibitem{memex}
T.~Malisiewicz and A.~Efros.
\newblock Beyond categories: The visual memex model for reasoning about object
  relationships.
\newblock {\em NIPS}, 2009.

\bibitem{DiMassa06}
V.~D. Massa, G.~Monfardini, L.~Sarti, F.~Scarselli, M.~Maggini, and M.~Gori.
\newblock A comparison between recursive neural networks and graph neural
  networks.
\newblock {\em IEEE International Joint Conference on Neural Network
  Proceedings}, 2006.

\bibitem{Micheli09}
A.~Micheli.
\newblock Neural network for graphs: A contextual constructive approach.
\newblock {\em IEEE Transactions on Neural Networks}, 2009.

\bibitem{Miller95}
G.~A. Miller.
\newblock Wordnet: A lexical database for english.
\newblock {\em ACM}, 38, 1995.

\bibitem{MisraNoisy16}
I.~Misra, C.~L. Zitnick, M.~Mitchell, and R.~Girshick.
\newblock {Seeing through the Human Reporting Bias: Visual Classifiers from
  Noisy Human-Centric Labels}.
\newblock In {\em CVPR}, 2016.

\bibitem{Niepert16}
M.~Niepert, M.~Ahmed, and K.~Kutzkov.
\newblock Learning convolutional neural networks for graphs.
\newblock {\em arXiv preprint arXiv:1605.05273}, 2016.

\bibitem{Orsini15}
F.~Orsini, P.~Frasconi, and L.~D. Raedt.
\newblock Graph invariant kernels.
\newblock {\em IJCAI}, 2015.

\bibitem{Vishwanathan15}
Pinar, Yanardag, and S.~V.~N. Vishwanathan.
\newblock Deep graph kernels.
\newblock {\em KDDM}, 2015.

\bibitem{renNIPS15fasterrcnn}
S.~Ren, K.~He, R.~Girshick, and J.~Sun.
\newblock Faster {R-CNN}: Towards real-time object detection with region
  proposal networks.
\newblock {\em NIPS}, 2015.

\bibitem{ILSVRC15}
O.~Russakovsky, J.~Deng, H.~Su, J.~Krause, S.~Satheesh, S.~Ma, Z.~Huang,
  A.~Karpathy, A.~Khosla, M.~Bernstein, A.~C. Berg, and L.~Fei-Fei.
\newblock {ImageNet Large Scale Visual Recognition Challenge}.
\newblock {\em IJCV}, 115(3):211--252, 2015.

\bibitem{Viske}
F.~Sadeghi, S.~K. Divvala, and A.~Farhadi.
\newblock Viske: Visual knowledge extraction and question answering by visual
  verification of relation phrases.
\newblock {\em CVPR}, 2015.

\bibitem{Scarselli09}
F.~Scarselli, M.~Gori, A.~C. Tsoi, and G.~Monfardini.
\newblock The graph neural network model.
\newblock {\em IEEE Transactions on Neural Networks}, 2009.

\bibitem{Shervashidze11}
N.~Shervashidze, P.~Schweitzer, E.~J. van Leeuwen, K.~Mehlhorn, and K.~M.
  Borgwardt.
\newblock Weisfeiler-lehman graph kernels.
\newblock {\em JMLR}, 2011.

\bibitem{Shervashidze09}
N.~Shervashidze, S.~V.~N. Vishwanathan, T.~H. Petri, K.~Mehlhorn, and K.~M.
  Borgwardt.
\newblock Efficient graphlet kernels for large graph comparison.
\newblock {\em AISTATS}, 5, 2009.

\bibitem{Shrivastava14}
A.~Shrivastava, S.~Singh, and A.~Gupta.
\newblock Constrained semi-supervised learning using attributes and comparative
  attributes.
\newblock {\em ECCV}, 2012.

\bibitem{VGG14}
K.~Simonyan and A.~Zisserman.
\newblock Very deep convolutional networks for large-scale image recognition.
\newblock {\em arXiv preprint arXiv:1409.1556}, 2014.

\bibitem{Vishwanathan10}
S.~V.~N. Vishwanathan, N.~N. Schraudolph, R.~Kondor, and K.~M. Borgwardt.
\newblock Graph kernels.
\newblock {\em JMLR}, 2010.

\bibitem{Zhu14}
X.~Zhu, D.~Anguelov, and D.~Ramanan.
\newblock Capturing long-tail distributions of object subcategories.
\newblock {\em CVPR}, 2014.

\bibitem{zhu2014eccv}
Y.~Zhu, A.~Fathi, and L.~Fei-Fei.
\newblock {Reasoning about Object Affordances in a Knowledge Base
  Representation}.
\newblock In {\em {European Conference on Computer Vision}}, 2014.

\bibitem{Zhu15}
Y.~Zhu, C.~Zhang, C.~Ré, and L.~Fei-Fei.
\newblock Building a large-scale multimodal knowledge base system for answering
  visual queries.
\newblock {\em arXiv preprint arXiv:1507.05670}, 2015.

\end{thebibliography}
}

\end{document}